\documentclass[10pt,twocolumn,letterpaper]{article}

\usepackage{iccv}
\usepackage{times}
\usepackage{epsfig}
\usepackage{graphicx}
\usepackage{amsmath}
\usepackage{amssymb}
\usepackage{multirow}
\usepackage{float}

\usepackage[pagebackref=true,breaklinks=true,letterpaper=true,colorlinks,bookmarks=false]{hyperref}

\iccvfinalcopy 

\def\iccvPaperID{****} 

\ificcvfinal\fi

\begin{document}


\title{1st Place Solution for PSG competition with ECCV'22 SenseHuman Workshop}

\author{Qixun Wang, Xiaofeng Guo, Haofan Wang\\
Xiaohongshu Inc; Institute of Optics and Electronics, Chinese Academy of Sciences\\
{\tt\small wangqixun@xiaohongshu.com, xiaofengguo2010@gmail.com, wanghaofan@xiaohongshu.com}
}

\maketitle
\ificcvfinal\thispagestyle{empty}\fi


\begin{abstract}

\textbf{P}anoptic \textbf{S}cene \textbf{G}raph (PSG) generation aims to generate scene graph representations based on panoptic segmentation instead of rigid bounding boxes. Existing PSG methods utilize one-stage paradigm which simultaneously generates scene graphs and predicts semantic segmentation masks or two-stage paradigm that first adopt an off-the-shelf panoptic segmentor, then pairwise relationship prediction between these predicted objects. 
One-stage approach despite having a simplified training paradigm, its segmentation results are usually under-satisfactory, while two-stage approach lacks global context and leads to low performance on relation prediction.
To bridge this gap, in this paper, we propose \textbf{GRNet}, a Global Relation Network in two-stage paradigm, where the pre-extracted local object features and their corresponding masks are fed into a transformer with class embeddings. To handle relation ambiguity and predicate classification bias caused by long-tailed distribution, we formulate relation prediction in the second stage as a multi-class classification task with soft label.
We conduct comprehensive experiments on OpenPSG dataset and achieve the state-of-art performance on the leadboard\footnote{https://www.cvmart.net/race/10349/base}. We also show the effectiveness of our soft label strategy for long-tailed classes in ablation studies. Our code has been released in \url{https://github.com/wangqixun/mfpsg}.

\end{abstract}


\section{Introduction}

Scene graph generation (SGG) task~\cite{johnson2015image} typically aims to generate comprehensive and structured representations for localized objects grounded by bounding boxes together with their pairwise relationships, which is critical for scene understanding in images. However, such a bounding box-based paradigm from a detection perspective suffers from several issues. First, only a coarse localization of objects are provided and there exists ambiguous pixels belonging to different objects or categories. Second, context regions such as background usually cannot be covered by bounding boxes. Third, there are inconsistent and redundant labels in SGG datasets. To address these limitations, \cite{yang2022panoptic} proposes the new task, panoptic scene graph generation (PSG) along with a large dataset OpenPSG\footnote{https://psgdataset.org} based on COCO~\cite{lin2014microsoft} and Visual Genome (VG)~\cite{krishna2017visual}, to facilitate the progress of scene graph representations based panoptic segmentations instead of rigid bounding boxes. Existing PSG methods tackle the problem in one-stage paradigm or two-stage paradigm. IMP~\cite{xu2017scene}, MOTIFS~\cite{zellers2018neural}, VC-Tree~\cite{tang2019learning}, and GPSNet~\cite{lin2020gps} are from the SGG task and are adapted for the PSG task as two-stage baselines. Recently, \cite{yang2022panoptic} designs a one-stage method named PSGTR based on the end-to-end DETR~\cite{carion2020end} to predict triples and localizations simultaneously. PSGFormer~\cite{yang2022panoptic} is further proposed on the top of PSGTR with an explicit relation modeling with a prompting-like matching mechanism and has shown competitive performance.

\begin{figure}
\centering
\includegraphics[width=1\linewidth]{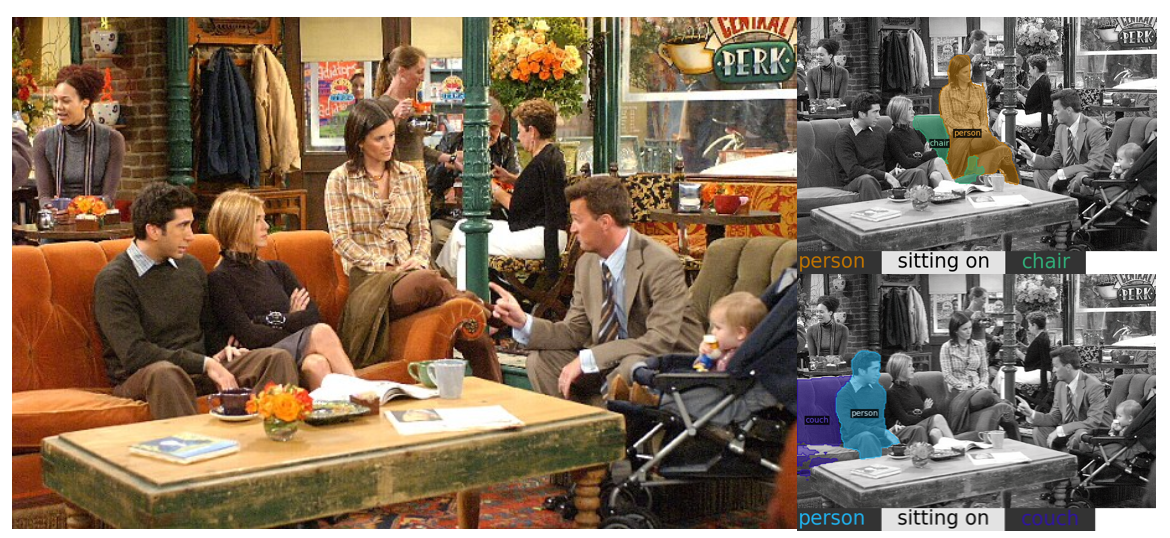}
\vspace{-3mm}
\caption{\textbf{Panoptic Scene graph Generation}. PSG presents a more comprehensive and clean scene graph representation, with more accurate localization of objects and including relationships with the background (known as stuff).}
\label{model}
\vspace{-3mm}
\end{figure}

Despite of the great potential of one-stage paradigm, we claim that the choice of one-stage and two-stage is still an open question for PSG task given the fact that two-stage paradigm still dominates SSG task. We analyze these two paradigms and find that one-stage methods benefit from global context and lead to better relation prediction performance, but suffer from unsatisfactory segmentation and instable training. While two-stage methods disentangle segmentation and relation prediction, thus have better segmentation results, nevertheless, they only extract local region features without global context and are not suitable for predicting relation between two distance objects. The trade-off between these paradigm inspires us to bridge such a gap.

In this work, we revive the two-stage paradigm by equipping it with ability to obtain global context as one-stage paradigm. Specifically, we first adopt Mask2Former~\cite{cheng2021mask2former} as our off-the-shelf panoptic segmentor, which generates masks for each object. The intermediate feature map of a specific object from segmentor and its corresponding mask are fused as object-level features. Instead of handling pairwise objects individually, we propose to build the global context utilizing a transformer that is fed with all found objects. A class embedding is added to indicate object class. Therefore, the $N \times N$ relations are jointly optimized in the training process, where $N$ is the number of objects.

Furthermore, we find that the performance of existing methods are also severely limited by issues associated with dataset. In concise, the relation ambiguity and inevitable bias caused by long-tailed distribution. To handle these, we re-formulate the relation prediction in second stage from traditional classification task to multi-class classification~\cite{tsoumakas2007multi} supervised with self-distilled soft label\cite{li2021self} instead of one-hot label. This also alleviates the negative effect of missing label in the dataset. Meanwhile, a focal loss~\cite{lin2017focal} is also adopted to re-balance loss from rare categories.

We evaluate our proposed framework on OpenPSG dataset following the default setting in ~\cite{yang2022panoptic} and conduct ablation studies to show the effectiveness of self-distilled soft label and focal loss to handle dataset associated issues. We claim that \textbf{GRNet} achieves the-state-of-art performance on the leaderboard and exceeds previous methods by a large margin. We will release the code upon acceptance.

In summary, our contributions are from three-fold:

\begin{itemize}
\item  We propose a new method called GRNet with a global relation module to bridge the gap between existing two-stage and one-stage methods.

\item  We formulate the relation prediction as a multi-class classification with soft label, which effectively alleviates relation ambiguity and long-tailed issues.

\item  We report the state-of-art performance on OpenPSG leaderboard, and exceed previous both two-stage and one-stage methods by a large margin.
\end{itemize}

\section{Methodology}

\begin{figure*}
\centering
\includegraphics[width=1\linewidth]{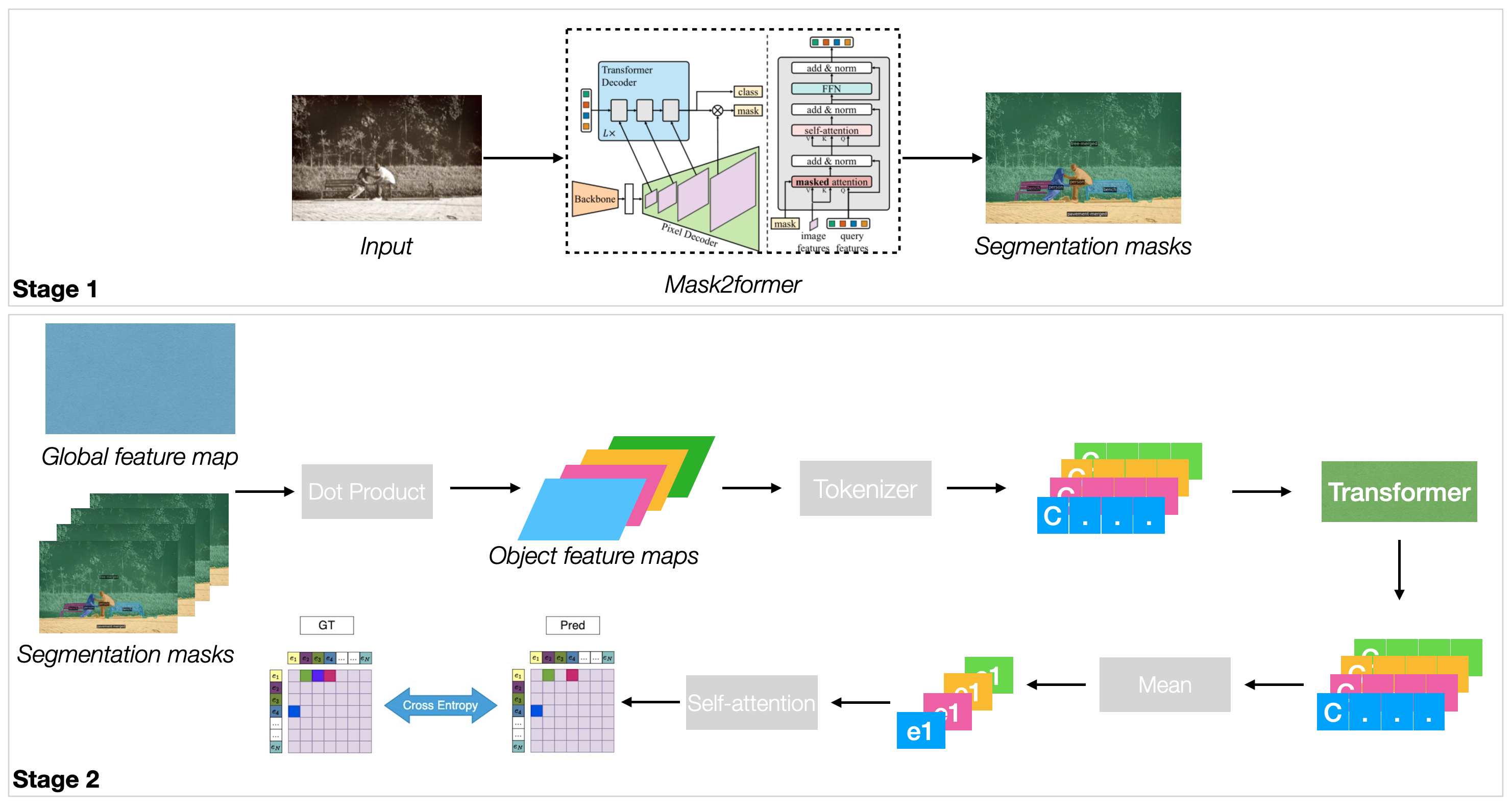}
\vspace{-3mm}
\caption{\textbf{The pipeline of our proposed GRNet}. We follow two-stage paradigm where an off-theshelf panoptic segmentor is used to generate object mask and then pairwise relationship prediction between these predicted objects.}
\label{fig_pipeline}
\vspace{-3mm}
\end{figure*}

Existing approaches tackle the problem in one stage or two stages. For one-stage methods, they are designed to predict relation triples and masks simultaneously. While two-stage methods usually adopt an off-the-shelf mask predictor first, then
pairwise relationship prediction between these predicted objects. 
However, there is still a clear trade-off between prediction performance and global relation construction. In this section, we first briefly recap the formulation of PSG task, and introduce our two-stage approach with the ability to construct global relation. To handle relation ambiguity and missing labels, we further propose self-distillation to produce soft labels.

\subsection{Problem Formulation}

The panoptic scene graph generation (PSG) task aims to model the following distribution:

\begin{equation}
Pr(G | I) = Pr(M,O,R | I)
\end{equation}

where $I\in R^{H\times W\times 3}$ is the given input image, and $G$ is desired scene graph which comprises the object masks $M = \{m_1, m_2, ..., m_n\}$ and class labels $O = \{o_1, o_2, ..., o_n\}$ that correspond to each of the $n$ objects in the image, and their relations in the set $R = \{r_1, r_2, ..., r_l\}$. Each mask is associated with an object with class label $o_{i}$. The masks do not overlap. More specifically, $m_i \in \{0,1\}^{H\times W}$ represents the binary mask, $o_i$ and $r_i$ belong to the set of all object and relation classes. For conciseness, both objects and background are referred as objects.

\subsection{Global Context Modeling}

\begin{figure}
\centering
\includegraphics[width=1\linewidth]{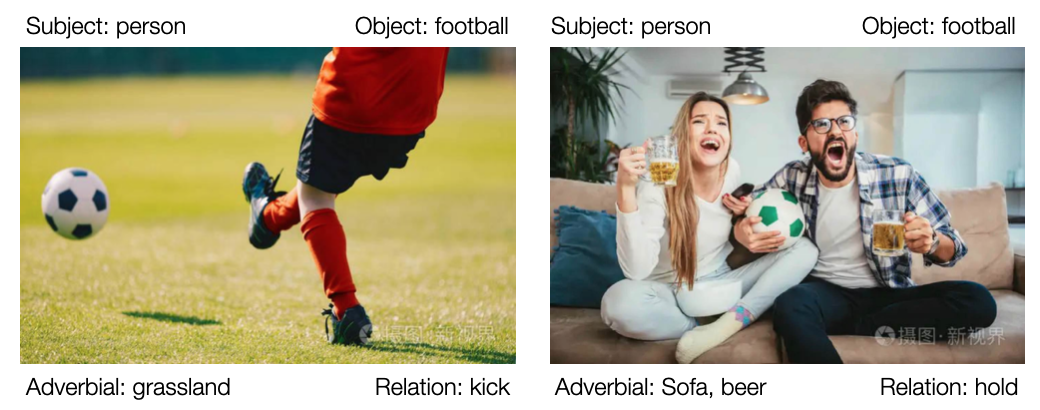}
\vspace{-3mm}
\caption{Illustration of the necessity of global contextual information (non-object entities) for relation prediction.}
\label{fig_32}
\vspace{-3mm}
\end{figure}

\textbf{Context Matters.} 
Previous two-stage methods only build pairwise relations between subjects and objects in a triplet, while ignoring the influence of rich global context from other neighborhoods. Let's take Figure \ref{fig_32} as an example, the subject and object are ``person" and ``football" respectively. However, their relationship varies from context to context. When there are adverbials of grassland, the predicate is likely to be the relationship of ``kicking". When there are adverbials of sofa, beer, and TV, the predicate is likely to be ``holding". Therefore, we claim that contextual information from non-object entities are also significant for predicting complex scene graph.

\textbf{Object Feature Tokens.} 
In our work, instead of pairwise subject and object, we construct global relation among uncertain entities via sequences modeling. The structure of our global context module is shown in Figure. Specifically, given feature map $f \in R^{H\times W}$ and a set of masks $M$ from our off-the-shelf mask predictor, we first obtain object-level feature $f_{i} = f \cdot m_i$, where $\cdot$ denotes dot product. As $f_i$ is already in hidden space, we do not use extra linear project layer. To tokenize each object feature, $f_{i}$ is flatten and divided into $L$ patches. Each patch is then applied with an average pooling to obtain its token. In order to perform classification, we use the standard approach of adding an extra learnable ``classification token'' to the token sequence. 

\textbf{Global Context Module.} 
After tokenization, we obtain $N \times (L+1)$ tokens that represent all objects within the given image, where $N$ is the count of objects. The token sequence is fed into a transformer to conduct $N \times N$ global modeling. Finally, for each object-level feature, we take global average pooling to obtain $N$ object embeddings with contextual enhancement. The message passing is achieved over all neighborhoods that makes use of the edge direction information and global contextual information.

\subsection{Multi-label Classification}

\textbf{Predicates Ambiguity.} 
The rigorous annotation process is conducted in the PSG task to ensure precision and salience, where the annotators strictly not to annotate using general relations like ``on" and ``in" when a more precise predicate like ``parked on" is applicable, with the assumption that there exists only a single relationship for a particular object pair. However, it is usually invalid in common phenomenon that a pair of objects can be attached by multiple relationships. In other words, although the different relationships are hierarchical, they are not mutually exclusive. We term this problem as predicate ambiguity. At the same time, we believe that this labeling ambiguity will make it difficult for the model to converge. A natural solution is that we convert the relationship prediction problem from the original one-label classification problem to a multi-label classification problem.

\textbf{Self-Distillation with Soft labels.}
Due to the task properties, there is only a more accurate relation annotation for each paired object. To generate multiple labels in practise, we lean on self-distillation to generate soft labels in the training process. To be more specific, we make use of the exponential moving average (EMA) to dynamic update our self-distilled predictor. Given an initialized relation predictor $Pr$, we first regularly train it with the hard labels at the beginning epochs until it converges to a local minima. Then, we set $Pr_{ema} = Pr$ and update our $Pr_{ema}$ as following, where $\alpha$ is the weight decay.

\begin{equation}
Pr_{ema} = Pr_{ema} * \alpha + Pr * (1-\alpha)
\end{equation}

We use $Pr_{ema}$ as our pseudo-label generator to obtain multi-labels on the training samples and finetune on soft labels for several epochs. We adopt multi-head attention layers in $Pr$ to construct relations between different entities, and the number of head is set to the number of relations.





\begin{table*}[]
\centering
\begin{tabular}{|c|c|c|c|ccc|}
\hline
                            &                             &                                &                          & \multicolumn{3}{c|}{Scene Graph Generation}                                                                                                      \\ \cline{5-7} 
\multirow{-2}{*}{Backbone}  & \multirow{-2}{*}{Paradigm}  & \multirow{-2}{*}{Segmentor}    & \multirow{-2}{*}{Method} & \multicolumn{1}{c|}{R/mR@20}                          & \multicolumn{1}{c|}{R/mR@50}                          & R/mR@100                         \\ \hline
                            &                             &                                & IMP                      & \multicolumn{1}{c|}{16.5/6.52}                        & \multicolumn{1}{c|}{18.2/7.05}                        & 18.6/7.23                        \\ \cline{4-7} 
                            &                             &                                & MOTIFS                   & \multicolumn{1}{c|}{20.0/9.10}                        & \multicolumn{1}{c|}{21.7/9.57}                        & 22.0/9.69                        \\ \cline{4-7} 
                            &                             &                                & VCTree                   & \multicolumn{1}{c|}{20.6/9.70}                        & \multicolumn{1}{c|}{22.1/10.2}                        & 22.5/10.2                        \\ \cline{4-7} 
                            &                             &                                & GPSNet                   & \multicolumn{1}{c|}{17.8/7.03}                        & \multicolumn{1}{c|}{19.6/7.49}                        & 20.1/7.67                        \\ \cline{4-7} 
                            &                             & \multirow{-5}{*}{Panoptic-FPN} & \textbf{GRNet (Ours)}    & \multicolumn{1}{c|}{31.0/25.9}                        & \multicolumn{1}{c|}{37.2/34.0}                        & 40.3/37.5                        \\ \cline{3-7} 
                            & \multirow{-6}{*}{Two-Stage} & Mask2former                    & \textbf{GRNet (Ours)}    & \multicolumn{1}{c|}{38.1/32.5}                        & \multicolumn{1}{c|}{46.2/41.7}                        & 50.3/48.9                        \\ \cline{2-7} 
                            &                             & -                              & PGSTR                    & \multicolumn{1}{c|}{28.4/16.6}                        & \multicolumn{1}{c|}{34.4/20.8}                        & 36.3/22.1                        \\ \cline{3-7} 
\multirow{-8}{*}{ResNet-50} & \multirow{-2}{*}{One-Stage} & -                              & PSGFormer                & \multicolumn{1}{c|}{18.0/14.8}                        & \multicolumn{1}{c|}{19.6/17.0}                        & 20.1/17.6                        \\ \hline
Swin-T                      & Two-Stage                   & Mask2former                    & \textbf{GRNet (Ours)}    & \multicolumn{1}{c|}{{39.2/31.9}} & \multicolumn{1}{c|}{{ 47.7/43.2}} & {51.9/49.5} \\ \hline
Swin-B                      & Two-Stage                   & Mask2former                    & \textbf{GRNet (Ours)}    & \multicolumn{1}{c|}{{40.8/34.2}} & \multicolumn{1}{c|}{{50.2/45.8}} & {55.0/52.1} \\ \hline
\end{tabular}
\vspace*{3mm}
\caption{Comparison between all baselines and our GCNet. Recall (R) and mean recall (mR) are reported.}
\label{table3}
\end{table*}

\section{Experiments}

In this section, we first introduce the dataset and the evaluation protocol for the PSG task in Section \ref{Dataset and Metric}. Implementation details are available in Section \ref{Implementation Details}. The comparisons with state-of-the-arts are provided in Section \ref{Comparisons with State-of-the-Arts}. 


\subsection{Dataset and Metric}
\label{Dataset and Metric}
\textbf{Dataset.} The OpenPSG dataset has 48,749 images with 133 object classes (80 objects and 53 stuff) and 56 predicate classes. It annotates inter-segment relations based on COCO panoptic segmentation.

\textbf{Metric.} PSG task comprises two sub-tasks: predicate classification (when applicable) and scene graph generation (main task) to evaluate the PSG models. Predicate classification (PredCls) aims to generate a scene graph given the ground-truth object labels and localization. Scene graph generation (SGDet) aims to generate scene graphs from scratch, which is the main result for the PSG task.

The classic metrics of R@K and mR@K are used to evaluate the previous two
sub-tasks, which calculates the triplet recall and mean recall for every predicate category, given the top K triplets from the PSG model. Notice that PSG grounds objects with segmentation, a successful recall requires both subject and object to have mask-based IOU larger than 0.5 compared to their ground-truth counterparts, with the correct classification on every position in the S-V-O triplet. The panoptic segmentation evaluation protocol PQ can be also considered as an auxiliary metric.

\subsection{Implementation Details}
\label{Implementation Details}

We adopt Mask2former with Swin-B backbone from MMDetection\footnote{https://github.com/open-mmlab/mmdetection} to obtain object masks and extract their RoI features. Since the performance of two-stage paradigm highly depends on the underlying segmentor, we also use the same set of parameters (ResNet backbone and Panoptic FPN) as previous works for fair comparison. The parameters of the pre-trained Mask2former are not frozen during training. We first train Mask2former individually for 50 epochs, and then jointly train the full GRNet for extra 12 epochs. AdamW is adopted with lr=0.0001 and weight\_decay=0.05. The learning rate is decayed 10x at 6th and 10th epoch. The weight of self-attention layer is adopted from hfl/chinese-roberta-wwm-ext-large\footnote{https://huggingface.co/hfl/chinese-roberta-wwm-ext-large}.

\subsection{Comparisons with State-of-the-Arts}
\label{Comparisons with State-of-the-Arts}

\textbf{Comparing Methods.}
We select baselines in two different paradigms. For one-stage approaches, we compare with PSGTR and PSGFormer. For two-stage approaches, we compare with IMP, MOTIFS, VCTREE and GPSNet.

\textbf{Quantitative Analysis.}
We compared the proposed \textit{GCNet} with state-of-the-art methods in Table \ref{table3}. As shown, GRNet (ResNet-50, Panopitic-FPN) outputperforms all two-stage methods by large margin under the same footprint, which validates the effectiveness of our designed modules straightforwardly. We also show the impressive performance gain brought by better backbone (Swin-T and Swin-B) and segmentor (Mask2former).






\section{Conclusion}
In our submission to the PSG Challenge @ ECCV'22 SenseHuman Workshop, we propose GR-Net with global context to boost the performance of two-stage methods. We formulate relation prediction as a multi-class classification task with soft label to further handle long-tailed issues. We claim that GR-Net achieves the-state-of-art performance on the leaderboard and exceeds previous methods (both two-stages and one-stage) by a large margin.

\newpage

{\small
\bibliographystyle{ieee_fullname}
\bibliography{egbib}
}

\end{document}